\documentclass[letterpaper]{article} %
\PassOptionsToPackage{dvipsnames}{xcolor}
\usepackage[]{aaai2026}  %
\usepackage{times}  %
\usepackage{helvet}  %
\usepackage{courier}  %
\usepackage[hyphens]{url}  %
\usepackage{graphicx} %
\usepackage{natbib}  %
\usepackage{caption} %
\usepackage{newfloat}
\usepackage{listings}
\DeclareCaptionStyle{ruled}{labelfont=normalfont,labelsep=colon,strut=off} %
\usepackage{microtype}
\usepackage{graphicx}
\usepackage{subfigure}
\usepackage{booktabs} %

\usepackage{minted}

\usepackage{tcolorbox}

\RequirePackage{graphicx}
 \usepackage{booktabs}
\usepackage{longtable}%
\makeatletter
\def\set@curr@file#1{\def\@curr@file{#1}} %
\makeatother
\usepackage[]{siunitx} %

\usepackage[linesnumbered,ruled,vlined,noend]{algorithm2e}

 \usepackage{adjustbox}

\usepackage{todonotes}
\usepackage{physics}
\usepackage{amsmath}
\usepackage{tikz}
\usepackage{mathdots}
\usepackage{cancel}
\usepackage{color}
\usepackage{array}
\usepackage{multirow}
\usepackage{gensymb}
\usepackage{extarrows}
\usepackage{booktabs}
\usetikzlibrary{fadings}
\usetikzlibrary{patterns}
\usetikzlibrary{shadows.blur}
\usetikzlibrary{shapes}

\usepackage{physics}
\usepackage{amsmath}
\usepackage{tikz}
\usepackage{mathdots}
\usepackage{yhmath}
\usepackage{cancel}
\usepackage{color}
\usepackage{array}
\usepackage{multirow}
\usepackage{amssymb}
\usepackage{gensymb}
\usepackage{extarrows}
\usepackage{booktabs}
\usetikzlibrary{fadings}
\usetikzlibrary{patterns}
\usetikzlibrary{shadows.blur}
\usetikzlibrary{shapes}

\title{Bayesian Meta-Analyses Could Be More: A Case Study in Trial of Labor After a Cesarean-section Outcomes and Complications}
\author{
    Ashley Klein\textsuperscript{\rm 1}\equalcontrib,
    Edward Raff\textsuperscript{\rm 2,3}\equalcontrib,
    Marcia DesJardin\textsuperscript{\rm 4}
}
\affiliations{
    \textsuperscript{\rm 1}Loftus, Ryu and Bartol OBGYN\\
    \textsuperscript{2} CrowdStrike\\
    \textsuperscript{3} University of Maryland, Baltimore County\\
    \textsuperscript{4} The University of Alabama at Birmingham
}

\begin{document}

\maketitle

\begin{abstract}
The meta-analysis's utility is dependent on previous studies having accurately captured the variables of interest, but in medical studies, a key decision variable that impacts a physician's decisions was not captured. This results in an unknown effect size and unreliable conclusions. A Bayesian approach may allow analysis to determine if the claim of a positive effect is still warranted, and we build a Bayesian approach to this common medical scenario. To demonstrate its utility, we assist professional OBGYNs in evaluating Trial of Labor After a Cesarean-section (TOLAC) situations where few interventions are available for patients and find the support needed for physicians to advance patient care.  
\end{abstract}

\section{Introduction}

Meta-analysis is one of the most important and critical tools in medical research~\citep{Lee2018,Haidich2010-gt} used to collect results from prior studies and coalesce them into a single, more robust conclusion. Medical research broadly is dominated by the use of the two simplest kinds of meta-analysis, fixed and random effects~\citep{Reis2023,KHackenberger2020}, which are useful tools but assume that the underlying studies have all the needed details and context. This becomes problematic when under-served areas like women's health result in
systemic missing information in medical studies. Neither a random nor fixed effect model will reach a reliable conclusion in such a situation, and this results in the inadequate study of conditions that impact vast portions of the population. Lack of thorough and rigorous studies in obstetric care is unfortunately common due to a lack of representation of minority groups~\citep{Steinberg2023}, lack of statistical rigor and power in OBGYN studies~\citep{Bruno2022}, and the lowest rate of research funding~\citep{Parchem2022}. Existing meta-analysis techniques for dealing with confounding factors require specifying the size of the confounding effect~\citep{VanderWeele_Ding_2017}. Since the confounding effect size is often unknown, this becomes unfalsifiable and leaves medical researchers no path forward beyond an expensive re-study of the problem, which is challenging due to the aforementioned lack of funding~\citep{Parchem2022}. The question then becomes, how can we leverage prior studies and data, even when it may have missing confounders to be corrected for?

In this article, we argue and show that a Bayesian approach to performing a meta-analysis can be used to help mitigate these prior errors and provide evidence for alternative hypotheses. Specifically, we address a common scenario where a decision variable is not incorporated into the meta-analysis. We then present a meta-analysis approach that can be rapidly adapted to new situations and describe our collaboration with clinicians in setting the hyper-priors. If an effect is still detected, the process can proceed. If no effect is detected, the known missing variable should be collected in a new study, and no reliable evidence exists to counter the null hypothesis. 

This is critical in the medical field as it is unacceptable to begin a clinical trial without having a valid reason to conduct the trial in the first place due to managing the risks (i.e., to patients) and rewards (i.e., reasonable expectation of efficacy). 
Specifically, it may occur that the issue is not in the mechanical application of the statistics but that full consideration of the medical procedures, process, and decision-making may reveal causal considerations that would invalidate the result of a standard statistical analysis.  In our case study, we are concerned with the usage of Pitocin and mechanical dilation (a.k.a., a catheter) as methods of induction of labor for patients with a history of cesarean delivery. Prior work has identified Pitocin as a (allegedly) superior and safer method for achieving this goal, which is clinically relevant as a failed TOLAC has more complications than a scheduled elective repeat cesarean section (Cesarean section, further abbreviated as CS). However, this conclusion was drawn by performing standard statistical tests on observational data without accounting for how the choice to use Pitocin vs mechanical dilation is made in practice. By using a Bayesian approach to the meta-analysis design, we can correct for these missing factors. 

 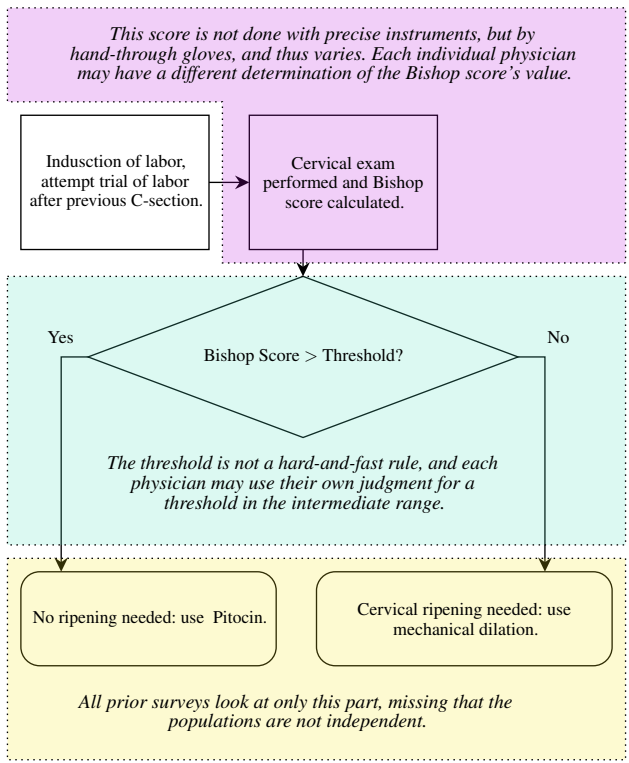
\begin{figure}[!h]
     \centering
     \adjustbox{max width=\columnwidth}{%
     \tikzset{every picture/.style={line width=0.75pt}} %

\begin{tikzpicture}[x=0.75pt,y=0.75pt,yscale=-1,xscale=1]
\draw   (230,210) -- (390,270) -- (230,330) -- (70,270) -- cycle ;
\draw   (20,444) .. controls (20,436.27) and (26.27,430) .. (34,430) -- (196,430) .. controls (203.73,430) and (210,436.27) .. (210,444) -- (210,486) .. controls (210,493.73) and (203.73,500) .. (196,500) -- (34,500) .. controls (26.27,500) and (20,493.73) .. (20,486) -- cycle ;
\draw   (240,444) .. controls (240,436.27) and (246.27,430) .. (254,430) -- (446,430) .. controls (453.73,430) and (460,436.27) .. (460,444) -- (460,486) .. controls (460,493.73) and (453.73,500) .. (446,500) -- (254,500) .. controls (246.27,500) and (240,493.73) .. (240,486) -- cycle ;
\draw   (20,90) -- (160,90) -- (160,190) -- (20,190) -- cycle ;
\draw   (190,90) -- (330,90) -- (330,190) -- (190,190) -- cycle ;
\draw    (160,140) -- (187,140) ;
\draw [shift={(190,140)}, rotate = 180] [fill={rgb, 255:red, 0; green, 0; blue, 0 }  ][line width=0.08]  [draw opacity=0] (10.72,-5.15) -- (0,0) -- (10.72,5.15) -- (7.12,0) -- cycle    ;
\draw    (230,190) -- (230,207) ;
\draw [shift={(230,210)}, rotate = 270] [fill={rgb, 255:red, 0; green, 0; blue, 0 }  ][line width=0.08]  [draw opacity=0] (10.72,-5.15) -- (0,0) -- (10.72,5.15) -- (7.12,0) -- cycle    ;
\draw    (390,270) -- (410,270) -- (410,427) ;
\draw [shift={(410,430)}, rotate = 270] [fill={rgb, 255:red, 0; green, 0; blue, 0 }  ][line width=0.08]  [draw opacity=0] (10.72,-5.15) -- (0,0) -- (10.72,5.15) -- (7.12,0) -- cycle    ;
\draw    (70,270) -- (50,270) -- (50,427) ;
\draw [shift={(50,430)}, rotate = 270] [fill={rgb, 255:red, 0; green, 0; blue, 0 }  ][line width=0.08]  [draw opacity=0] (10.72,-5.15) -- (0,0) -- (10.72,5.15) -- (7.12,0) -- cycle    ;
\draw  [fill={rgb, 255:red, 248; green, 231; blue, 28 }  ,fill opacity=0.2 ][dash pattern={on 0.84pt off 2.51pt}] (10,420) -- (470,420) -- (470,570) -- (10,570) -- cycle ;
\draw  [fill={rgb, 255:red, 189; green, 16; blue, 224 }  ,fill opacity=0.2 ][dash pattern={on 0.84pt off 2.51pt}] (470,200) -- (170,200) -- (170,80) -- (10,80) -- (10,10) -- (470,10) -- (470,200) -- cycle ;
\draw  [fill={rgb, 255:red, 80; green, 227; blue, 194 }  ,fill opacity=0.2 ][dash pattern={on 0.84pt off 2.51pt}] (10,210) -- (470,210) -- (470,410) -- (10,410) -- cycle ;

\draw (90,140) node   [align=left] {\begin{minipage}[lt]{95.2pt}\setlength\topsep{0pt}
\begin{center}
Indusction of labor, attempt trial of labor after previous C-section.
\end{center}

\end{minipage}};
\draw (260,140) node   [align=left] {\begin{minipage}[lt]{95.2pt}\setlength\topsep{0pt}
\begin{center}
Cervical exam performed and Bishop score calculated. 
\end{center}

\end{minipage}};
\draw (230,270) node   [align=left] {\begin{minipage}[lt]{129.2pt}\setlength\topsep{0pt}
\begin{center}
Bishop Score $>$ Threshold?
\end{center}

\end{minipage}};
\draw (350,465) node   [align=left] {\begin{minipage}[lt]{133.85pt}\setlength\topsep{0pt}
\begin{center}
Cervical ripening needed: use mechanical dilation.
\end{center}

\end{minipage}};
\draw (115,465) node   [align=left] {\begin{minipage}[lt]{129.2pt}\setlength\topsep{0pt}
\begin{center}
No ripening needed: use \ Pitocin.
\end{center}

\end{minipage}};
\draw (50,255) node   [align=left] {\begin{minipage}[lt]{27.2pt}\setlength\topsep{0pt}
\begin{center}
Yes
\end{center}

\end{minipage}};
\draw (420,255) node   [align=left] {\begin{minipage}[lt]{40.8pt}\setlength\topsep{0pt}
\begin{center}
No
\end{center}

\end{minipage}};
\draw (225,535) node   [align=left] {\begin{minipage}[lt]{251.6pt}\setlength\topsep{0pt}
\begin{center}
\textit{{\large All prior surveys look at only this part, missing that the populations are not independent.}}
\end{center}

\end{minipage}};
\draw (245,45) node   [align=left] {\begin{minipage}[lt]{292.4pt}\setlength\topsep{0pt}
\begin{center}
\textit{{\large This score is not done with precise instruments, but by hand-through gloves, and thus varies. Each individual physician may have a different determination of the Bishop score’s value.}}
\end{center}

\end{minipage}};
\draw (230,365) node   [align=left] {\begin{minipage}[lt]{217.6pt}\setlength\topsep{0pt}
\begin{center}
\textit{{\large The threshold is not a hard-and-fast rule, and each physician may use their own judgment for a threshold in the intermediate range.}}
\end{center}

\end{minipage}};

\end{tikzpicture}
     }
\caption{Cervical ripening is achieved with the application of Pitocin or mechanical dilation (a catheter). Whether one method has a higher rate of successful vaginal delivery after a C-section is of clinical relevance to the decision, as a failed TOLAC has higher rates of complications than a scheduled repeat C-section. The choice is made by computing a Bishop Score in clinical practice but creates a shared dependency (i.e., in the same Markov blanket), and this factor was ignored in previous studies. This creates an unaccounted causal impact between the two groups.}
\label{fig:bishopWhatIs}
 \end{figure}

To understand this issue, we present the decision process with additional context in Fig. \ref{fig:bishopWhatIs}. By hand, a medical provider will measure a \textit{Bishop score}~\citep{Bishop1964-jl}. The primary purpose of a Bishop score is to indicate the likelihood of a successful vaginal induction, whereas higher scores indicate a higher likelihood. This is thus directly causal to the target outcome. For example, a component of the Bishop score is dilation of the cervix (i.e., how open your cervix is), and a closed cervix can not physically pass a baby. 

If a patient had a sufficiently high Bishop score (e.g., 13 out of a maximum of 13 points), no cervical ripening is needed, and so neither Pitocin nor mechanical dilation would be used. However, Pitocin is prescribed only when the Bishop score is sufficiently high; thus, less ripening is needed. Ergo, prior studies that have shown Pitocin as being statistically better are failing to disentangle that \textit{a patient receiving Pitocin has already been determined to have a more favorable cervix, and thus closer to a successful delivery at the onset}. 

We turn to Bayesian machine learning and probabilistic programming to answer this clinical research question and to act as a case study on how incorporating Bayesian methods with physician guidance can help correct historical inequities. Using the prior knowledge of how the decision process is made in Fig. \ref{fig:bishopWhatIs}, we will develop a graphical model that accounts for the Bishop score as a hidden variable. Further, we can model the impact of this unobserved variable by using the Pitocin vs catheter populations against a truncated hidden variable, where the provider's discretion on a Bishop threshold informs the truncation of the hidden variable. \textit{This allows us to replace subjective approaches from the literature that require already knowing the impact of the confounder with more objective hyper-priors that can be set by clinical guidance and prior published evidence}. In doing so, we can leverage the incomplete data (Bishop scores have not been reported in prior studies) to determine that there is no adverse statistical difference between Pitocin and mechanical dilation when accounting for the unobserved Bishop scores. This conclusion is important for women's healthcare, enabling more options when adverse reactions occur, aligning with provider skill, and providing the justification to pursue an RCT for confirmation. All research was carried out under strict guidance and cooperation with professional OBGYNs to ensure clinical relevance and transition to clinical practice. 

The rest of our article is organized as follows. First, we will provide the detailed medical context of our study in  \S \ref{sec:medical_context}. Second, we will describe the Bayesian model we construct and additional tests that will be performed in  \S \ref{sec:methods}. The results of our methods and clinical relevance will be demonstrated in  \S \ref{sec:results} with further discussion in  \S \ref{sec:discussion}. Finally, we will end with a comparison to prior methods for meta-analysis with confounding factors in  \S \ref{sec:confounded_prior} and our conclusion in  \S \ref{sec:conclusion}.

 \section{Medical Context} \label{sec:medical_context}

Our approach is readily adaptable to stations where a single decision variable was not factored into the original analysis or included in the data.
Such single-variable deciders occur regularly due to differences in care due to provider discretion ~\citep{Cook2018} and the use of simple decision criteria ~\citep{Bae2014,Podgorelec2002} that are prevalent throughout healthcare as a means of reducing research findings into easy-to-remember actionable rules that can be applied in fast-moving clinical environments. 

Our case study is chosen in part because of its applicability to most women in the world. The rate of cesarean sections has been increasing globally from 12.1\% in 2000 to 21.1\% in 2015~\citep{boerma2018a,ja2021a}. In the United States, the rate of cesarean section in 2019 was 31.7\%~\citep{osterman2020a}. Cesarean sections have a higher rate of adverse consequences, including infection, hemorrhage, and increased risk of complications, such as abnormal placentation, in future pregnancies. Vaginal delivery is, therefore, the preferred method of delivery when clinically possible. Optimal methods for trial of labor after cesarean section (TOLAC) for vaginal birth after cesarean section (VBAC) have been under investigation as an alternative to repeat cesarean section in hopes of reducing complications. 

In 2018, the rate of VBAC in the United States was 13.3\%~\citep{osterman2020a}. Successful VBAC has fewer adverse outcomes than repeat cesarean sections and reduces the risk associated with multiple cesarean sections ~\citep{sc2015a,silver2006a,nisenblat2006a}.  Failed TOLAC, however, is associated with increased morbidity and mortality compared to an elective repeat cesarean section~\citep{hibbard2001a}. Uterine rupture results in the most significant increase of morbidity and mortality with TOLAC. The rate of uterine rupture with one prior low-transverse cesarean delivery is $<$1\%; however, it is an obstetrical emergency if it occurs~\citep{landon1010a,lydon-rochelle1010a}. This risk leads to policies by hospitals and obstetricians to forbid induction of labor for patients desiring TOLAC or to limit induction to Pitocin and artificial rupture of membranes (AROM). The increase in the number of TOLACs has prompted further interest in optimizing the clinical management of patients attempting a VBAC to improve the likelihood of success. 

Induction methods for TOLAC have recently been under investigation. In 2013, Jozwiak and Dodd attempted to compare methods of labor induction for TOLAC, but there was insufficient evidence to make a conclusion~\citep{west2017a}. The effect of prostaglandins and risk of uterine rupture during TOLAC has shown mixed results~\citep{landon1010a,lydon-rochelle1010a}. Multiple studies show that misoprostol significantly increases risk of uterine rupture~\citep{plaut1999a,aslan2004a}. The American College of Obstetricians and Gynecologists (ACOG) recommends that misoprostol should not be used for cervical ripening in patients with previous cesarean section or uterine surgery, but does not set clear guidelines on the use of other prostaglandins. Studies on the use of mechanical dilation for induction for TOLAC, such as a cervical ripening balloon and foley catheter, have also had mixed results~\citep{ravasia2000a,bujold2004a,hoffman2004a}. ~\citep{kehl2016a} published a meta-analysis in 2016 that looked at induction methods in TOLACs. They found mechanical dilation to be as effective as prostaglandins, with a lower risk for uterine rupture. However, the review was limited by number and quality of studies. Multiple studies have been published since in an attempt to evaluate the effectiveness of mechanical dilation. ~\citep{wingert2019a} published a meta-analysis in 2019 that evaluated clinical interventions impacting VBAC rates but included limited information on the impact of mechanical dilation. This study aims to evaluate the safety and outcomes of mechanical dilation for cervical ripening during TOLAC. 

\section{Methods} \label{sec:methods}

The data collection of our meta-analysis will follow standard clinical protocols used by physicians to perform such studies.
Multiple OBGYN physicians performed this step of the work to collect the data, specify all search terms, and perform all data filtering, to ensure that our data for the study is clinical meta-analysis data as collected by OBGYNs attempting to elucidate a clinical research question. 
A result of this work is the start of new clinical trials to test the efficacy without the decision threshold, and our results cover the literature at the time the initial study was completed. 

In  \S \ref{sec:our_method_details}, our approach to a Bayesian meta-analysis will be detailed. A set of distributions are chosen that are clinically relevant to how the Bishop score is calculated, and broadly applicable to other scenarios. The generative story will be presented in the general form, with comments denoting how the hyper-priors and variables were set for our case study. The standard fixed-effect meta-analysis will be covered in  \S \ref{sec:fixed_effect} and used in comparison to show how results change and in the analysis of outcomes that are not related to the Bishop score.

\subsection{Meta-Analysis Method} \label{sec:our_method_details}

By default, and when applicable, we used the standard fixed-effect model to perform a meta-analysis across studies~\citep{v-a}. We used the fixed effect model primarily due to the small number of studies that have looked at this problem. While a random-effects model is often preferred in medical studies~\citep{higgins2009a}, the small number of n=6 total studies makes the estimation of random-effects models unreliable, and so a fixed-effects model is generally preferred~\citep{higgins2009a,lin2020a}. We use the relative risk measure as the effect size of the intervention (mechanical dilation) since the outcome is binary (cesarean delivery versus vaginal delivery) and an event of regular occurrence~\citep{nakayama1998a}. The fixed-effect model (and all other results) were computed using PythonMeta. 

However, the above-cited papers recommend taking a Bayesian approach to better handle difficult low-sample size situations when possible. We do so for the relative risk calculations because the data strongly violates the heterogeneity assumption of the fixed-effect modeling. We also note the mediating factor that the studies included select the use of a catheter based on physician's clinically recommended discretion, often in regard to unreported Bishop scores. This means the standard hypothesis testing assumes $\mathbb{P}$(CS$|$intervention), when the actual data reflects a second unobserved conditional on the Bishop score, giving $\mathbb{P}$(CS$|$intervention, Bishop). 

\subsubsection{Specification of Priors}

For each prior that needed to be set, they were done so by having the machine learning researcher(s) in this study consult with the physician(s) involved. The goal was to make the priors as informative to the underlying true generative process, rather than convenience in statistical inference (hence, the use of a truncated normal distribution between populations). To develop a shared and agreed approach for setting these priors, the following question process was developed and asked in this order:
\begin{enumerate}
    \item Is this event recorded at a national, state-wide, or hospital database --- with aggregate statistics available? If yes, the largest available population rate was used to specify a prior. 
    \item Is there a maximum plausible range of values one might expect (e.g., is going from minimum to maximum Bishop score plausible?) in the most extreme cases? Another framing that was helpful was ``what is an impact that would be so large to shock you that it was so effective''. This helped with deciding between normal vs. heavy-tailed priors, and if any prior should have a limited range or scope. 
    \item What would you expect to see in normal clinical practice, and how consistent is that observation?
\end{enumerate}

The order in which we recommend these questions is based on the degree to which we found them useful in this study. Large-scale statistics are the most objective, and maximum ranges were useful in specifying the $\tau$ priors on the degree of impact on the population rates (lines 16 and 17). The third question was still useful, but not to the same degree. We found question three often resulted in ``it-depends'' responses and ambiguity, but was valuable in driving the discussion into \textit{why }``it depends''. This further discussion would eventually lead to evolution in the graphical model or other questions that would help derive a prior. 

We note that the Bishop prior and threshold, by their nature as the unobserved decision variable, would not benefit from more nuanced prior specification because there are no direct observations to benefit from appropriate scaling. For this reason, normal distributions were chosen for ease of inference. We also note that in Fig. \ref{fig:bishopWhatIs}, there are multiple sources of variation: the measurement error of the Bishop score, the provider discretion on the threshold to use, and the bias of individual providers on their own measurements\footnote{To clarify the difference in the latter two, note that the provider making the decisions may not be the same individual measuring the Bishop score. For example, a medical resident operating under the guidance of an attending physician will have two different entities involved rather than one.}. These are more effects than our model can account for because the different types of errors are outside the Markov blanket of the observed variables, and so they collapse into a smaller number of discrete items we can infer.

\subsubsection{Bayesian-Meta Analysis Method}

\begin{figure}[!h]
    \centering
    \adjustbox{width=\columnwidth}{%
    \includegraphics[]{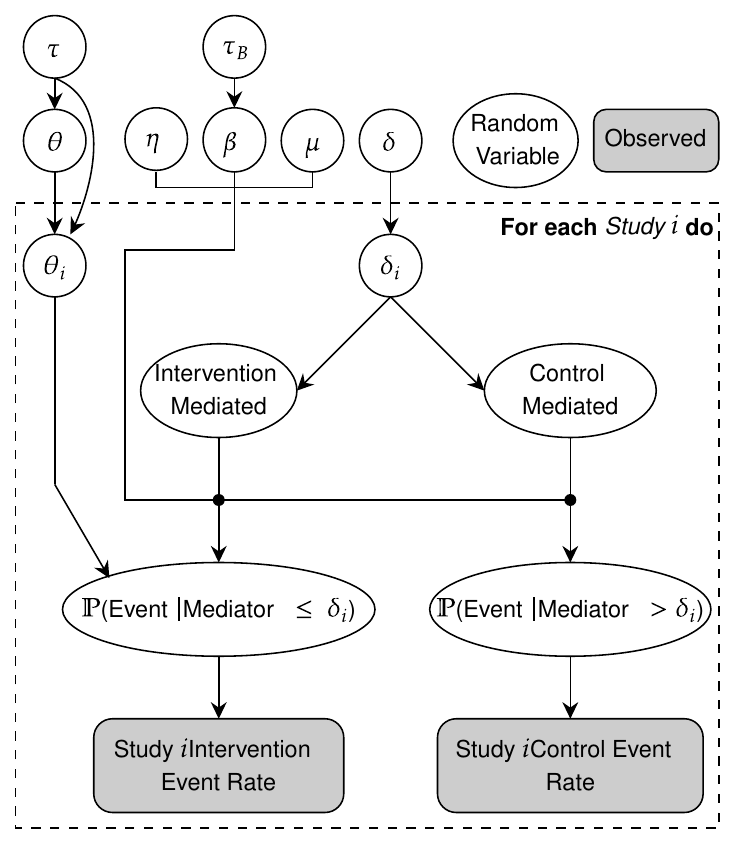}
    }
    \caption{The Plate Diagram for our model. Grey nodes are observed variables, white are sampled, and dashed lines are subgraphs that repeat together. The key observation is that there is a bifurcation of variables modeling 
    the outcome given a mediating decision variable (in this study, a Bishop score)
    for the control (Pitocin) and intervention (catheter). This shows how our algorithm accounts for a single mediator (Bishop score) and threshold (outside the num\_samples plate) that is altered by a provider's discretion $\delta_i$ and then causes the two groups to have different underlying Bishop scores.  }
    \label{fig:plateDiagram}
\end{figure}

We develop a general-purpose hierarchical Bayesian model to handle our situation: an unobserved (or otherwise not recorded) factor is used as a decision criterion, acting as a mediator on the apparent effectiveness of a control and an intervention on the occurrence of some adverse event $E$.  
The Plate diagram of our method is shown in Fig. \ref{fig:plateDiagram}, with key groups of variables. The $\delta$ variables model the unobserved decision variable, which will be left/right truncated for the intervention and control effects, respectively. $\beta$, $\mu$, and $\eta$ will control the base rate of occurrence in each population, which is mediated by the decision variable $\delta$. The $\theta$ variables will control the effect of the intervention. Finally, the $\tau$ variables are hyper-priors common to Bayesian meta-analyses. 

To perform a meta-analysis with our approach, only one hyper-prior \textit{needs} adjustment by the user for their own problems. A population-wide adverse-event rate $\mu$ must be specified as a weakly informative prior.  For our case study we use statistics from the US population rate of C-sections overall births in 2019 ~\citep{moher2015a}, which indicates a population-wide cesarean section rate of 31.7\%.
We will now detail how and why we set each of these priors to specific distributions, and their precise interactions, which are summarized in algorithm \ref{alg:generativeTolac}.

\begin{algorithm}[!h]
\SetAlgoLined
\caption{Generative story for decision mediated meta-analysis. Comments describe the purpose of variables, with the TOLAC example in parenthesis. }\label{alg:generativeTolac}
\KwData{Observed counts of the event (C-section) and non-event (vaginal birth) for each study segregated by the use of a catheter (mechanical dilation) for low Bishop scores and the use of Pitocin for the high Bishop scores.}
$\log(\eta) \sim \text{Logistic}(\log(30))$\; 

$\eta \gets \exp(\log(\eta))$\;

$\mu \sim \text{Beta}(12, 25)$ {\color{ForestGreen}\tcp*{Population-wide cesarean section rate prior}}

$\tau \sim \text{HalfCauchy}(\sqrt{0.5/3})$ %

$\tau_B \sim \text{HalfCauchy}(\sqrt{0.5/3})$ \;

$\theta \sim \mathcal{N}(0, \tau^2)$ {\color{ForestGreen}\tcp*{Impact of the intervention (mechanical dilation)}}

$\beta \sim \mathcal{N}(0, \tau_B^2)$ {\color{ForestGreen}\tcp*{Unobserved impact of the mediator (Bishop score) has on the outcome (needing a C-section)}}

$\mathit{MediatorPrior} \sim \mathcal{N}(0, 1)$ {\color{ForestGreen}\tcp*{Unobserved mediator (Bishop score)}}

$\delta \sim \mathcal{N}(0, 1)$ {\color{ForestGreen}{\color{ForestGreen}\tcp*{Mediator (Bishop) threshold location}}}
\ForEach{Study $i$}{%
    $\theta_i \sim \mathcal{N}(\theta,\tau^2)$

    $\mathit{MediatorDist} \gets \mathcal{N}(\mathit{MediatorPrior}, 1)$  %

    $\delta_i \sim \mathcal{N}(\delta, 1)$ {\color{ForestGreen}\tcp*{The study specific threshold that is inferred.}}

    $\mathit{Intervention Mediated} \sim \mathit{MediatorDist}_{(-\infty, \delta_i)}$ %

    $\mathit{Control Mediated} \sim \mathit{MediatorDist}_{[\delta_i, \infty)}$

    $\mu_{\mathit{control}} \gets \exp(\operatorname{logit}(\mu) + \beta \cdot \mathit{Control Mediated} )$

     $\mu_{\mathit{intervention}} \gets \exp(\operatorname{logit}(\mu) + \beta \cdot \mathit{Intervention Mediated} + \theta_i)$

    $p_{E}^{\uparrow} \sim \text{Beta}(\mu_{\mathit{control}} \cdot  \eta, (1-\mu_{\mathit{control}} ) \cdot  \eta)$ {\color{ForestGreen}\tcp*{ Probability of event in the control population ($\mathbb{P}(\text{C-section} | \text{High Bishop})$)}}

    $p_{E}^{\downarrow} \sim \text{Beta}(\mu_{\mathit{intervention}} \cdot \eta, (1-\mu_{\mathit{intervention}} )  \cdot  \eta)$ {\color{ForestGreen}\tcp*{ Probability of event in the intervention population ($\mathbb{P}(\text{C-section} | \text{Low Bishop})$)}}

    $RR \gets p_{E}^{\downarrow} / p_{E}^{\uparrow}$ {\color{ForestGreen}f\tcp*{Compute the Relative Risk}}
    Observe likelihoods of $p_{E}^{\downarrow}$ and $p_{E}^{\uparrow}$ against the observed study rate of event (C-sections). 
}
\end{algorithm}

We select a weakly informative prior of log($\eta$) $\sim $Logistic(log(30)) for the strength of the hyper prior of the Beta distributions that will be used for the probability of a C-section across studies. 
We model the intervention effect on the $\text{logit}(\mu) + \beta \cdot \operatorname{Truncated}+ \theta_i$ as a hyper prior, where $\operatorname{Truncated}$ indicates the control or mediator truncated portion of the (inferred latent) distribution.  $\theta \sim $Normal$(0, \tau^2)$ and study level prior $\theta_i \sim $Normal($\theta$, $\tau^2$), with $\tau \sim $HalfCauchy($\frac{\sqrt{0.5}}{3}$). The use of $\tau$ is dual purpose. First, it is recommended by ~\citep{harrer2021doing} for Bayesian meta-analysis to account for the between-study heterogeneity. Second, it also corresponds to a Horseshoe prior~\citep{Brennan_2021,Carvalho_Polson_Scott_2009} to correspond to a default of no-impact, which is easy for the model to overcome given such an observation~\citep{Bhadra_Datta_Polson_Willard_2019}. The scale is chosen so that at the logit scale, the prior impact of $\tau$ will be that the impact of the intervention should be within $\pm$ 50 percentage points, making it a weak prior informative to the likely range of impact based on OBGYN feedback. The $\theta_i$ is dropped from the control group (i.e., they have no intervention). 
Similarly, $\beta$ models the unobserved Bishop score, and $\beta \sim$ Normal(0, $\tau_B^2$) models the impact of the Bishop score on the logits of needing a CS ($\tau_B^2$ is modeled the same way as $\tau$ for the same reasons). 

As noted, we need to specify the problem-specific parameter $\mu$. For us, the population-wide C-section rate $\mu \sim$ Beta(12,25) is chosen correspond with an expectation that $\mu= 12/(12+25) = 32\% \approx 31.7\%$. For any other problem, the user can use $\mu \sim $Beta($P$, $N$) where $P$ and $N$ are positive/negative occurrences, respectively, such that $P/(P+N)$ approximates a known population-wide occurrence rate. A reasonable heuristic when using a known population rate $X$ is to set $P=X*50$, $N=(1-X)*50$. 

To account for the Bishop score being used in determining if a catheter is needed, we model the effect as Truncated distributions. The $\mathit{MediatorDist}$ represents the hidden distribution of Bishop scores, which is truncated by a low and high range (denoted by a subscript of [low,high]). In this manner, the fundamental decision process as outlined in Fig. \ref{fig:bishopWhatIs} is incorporated into the model and allowed to have an impact itself. %

The global Bishop score is modeled with a prior of Normal(0, 1) as it is unobserved, and our interest is in the impact on the model, making the scale irrelevant. The threshold of provider preference on Bishop scores is hierarchical with $\delta \sim $Normal(0, 1) as and provider level preference $\delta_i \sim $Normal($\delta$, 1). The values $\mathit{InterventionMediated}_i$ and $\mathit{ControlMediated}_i$ are then sampled from a lower/upper truncated (at $\delta_i$) Normal distribution reflecting the lower/higher Bishop scores in the catheter/control groups respectively (performed on lines 14 and 15). These hyperpriors provide valuable prior information to enable the possibility of reaching a conclusion despite few studies being available and provide population-wide estimates. 
As the model we choose is hierarchical to account for the potential heterogeneity in studies, we model each study's baseline rate of cesarean section as sampled from Beta($\mu \cdot \eta$,$(1-\mu)\cdot \eta$)). 

Our model  algorithm \ref{alg:generativeTolac} is computed using Markov Chain Monte Carlo~\citep{metropolis1953a} using the NUTS sampler~\citep{hoffman2014a} and implemented with Numpyro~\citep{phan-a}
with highest density regions (HDR) of 95\%. 

\subsection{Standard Fixed-Effect Models for Other Cases} \label{sec:fixed_effect}

In the medical context of this work, we also wish to compare the risk of adverse uterine and neonatal events, so we calculated the odds ratio. These variables do not have the same bias with Bishop scores, and so a standard fixed-effect model is desirable for both familiarity and appropriateness given the context. Both of these events are rare, often with 0-5 occurrences in either group.  This makes relative risk infeasible to estimate, and so the odds ratio is preferred under the ``rare-disease assumption''~\citep{Greenland1982OnTN,Cummings2001OnTN,greenland1986a}.  Even still, the odds ratio of significantly rare events is error prone, especially when no occurrences are detected in the control or intervention group~\citep{chang2017a}. We choose to use Peto’s method of estimating the approximate odds ratio of rare events in a fixed-effect meta-analysis~\citep{yusuf1985a}. Peto’s method has received significant study and has been found to be sufficient so long as the event's rareness is on the order of $\leq$ 1\% and the control/intervention groups are not lopsided in sample sizes, which our data satisfies~\citep{greenland1990a,brockhaus2014a,brockwell2001a}. While a similar Bayesian treatment is possible, the extremely limited number of occurrences makes the impact of uninformative priors let alone weakly informative, especially large and nuanced\citep{unknown2018a}. Such nuanced statistical modeling is beyond the scope of study. Thus, a well-studied and valid fixed-effect model is preferred. 

\section{Results} \label{sec:results}

	4037 patients that underwent TOLAC, were identified~\citep{bullough2021a,eriksen2019a,koenigbauer2021a,radan2017a,ralph2020a,sarreau2020a} The patients had a mean age of 31.46 years and a mean gestational age of 39.22 weeks. These patients were divided into a cohort of patients who underwent TOLAC with the use of mechanical dilation (n = 1039 patients) and a cohort who did not receive mechanical dilation (n = 2998 patients). The cohort of patients using mechanical dilation included those who used a foley balloon, Cook catheter, and osmotic dilators. 231 patients had a prior vaginal delivery in this group. In the studies that included the indications for prior cesarean delivery, these included labor dystocia (29\%), non-labor dystocia (i.e., fetal heart tracing abnormalities or fetal malpresentation; 50.6\%), and unspecified (21\%).\citep{brockhaus2014a,bullough2021a,koenigbauer2021a}

	The first outcome we consider is the relative risk of a cesarean delivery under intervention (mechanical dilation) compared to a control group (pitocin, AROM, dinoprostone). The vaginal delivery rate in the mechanical dilation group was 51.4\% compared to the control group 69.1\%. 

  \begin{figure}[!h]
    \centering
    \adjustbox{max width=\columnwidth}{%
    \includegraphics{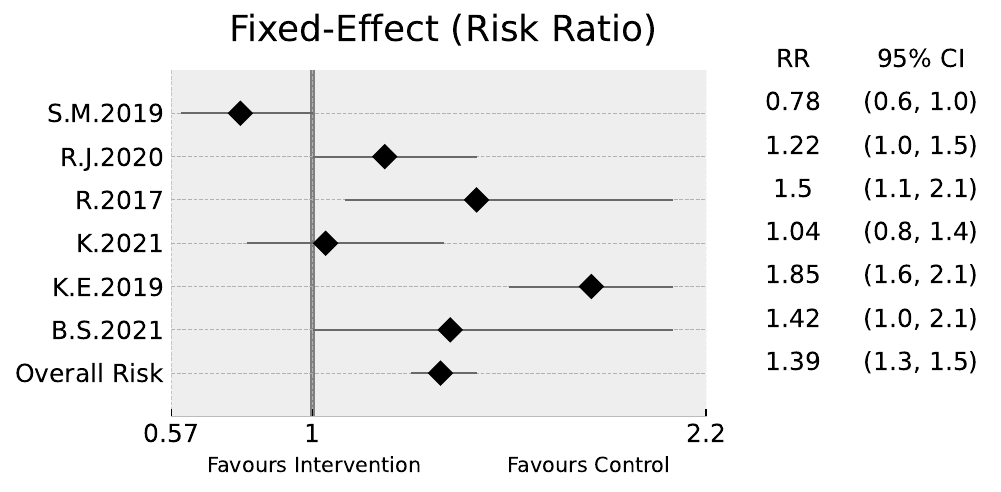}
    }
    \caption{Forest Plot of mechanical dilation (intervention) vs. Pitocin when ignoring the unaccounted Bishop score as outlined in Fig. \ref{fig:bishopWhatIs}. Because a Fixed-Effect meta-analysis fails to account for this mediating variable, and fails assumption tests, we do not recommend using its conclusion. It is presented to show why a Bayesian approach with a hidden variable for the Bishop score is needed.}
    \label{fig:fixedNaive}
\end{figure}
 
 Using a standard fixed-effect model (which ignores the causal impact of Bishop scores), the results are significant with p$<$0.001 and a mean cesarean section estimated relative risk of 1.39 (CI 1.27-1.51), indicating the risk of cesarean section in the mechanical dilation group is 39\% higher (Fig. \ref{fig:fixedNaive}), but a Cochran’s Q test shows significant heterogeneity (p$<$0.001) that, combined with the impact of mechanical dilation selected by physician preference, leads us to be doubtful of this result. 
 
\begin{figure}[!h]
    \centering
    \adjustbox{max width=\columnwidth}{%
    \includegraphics{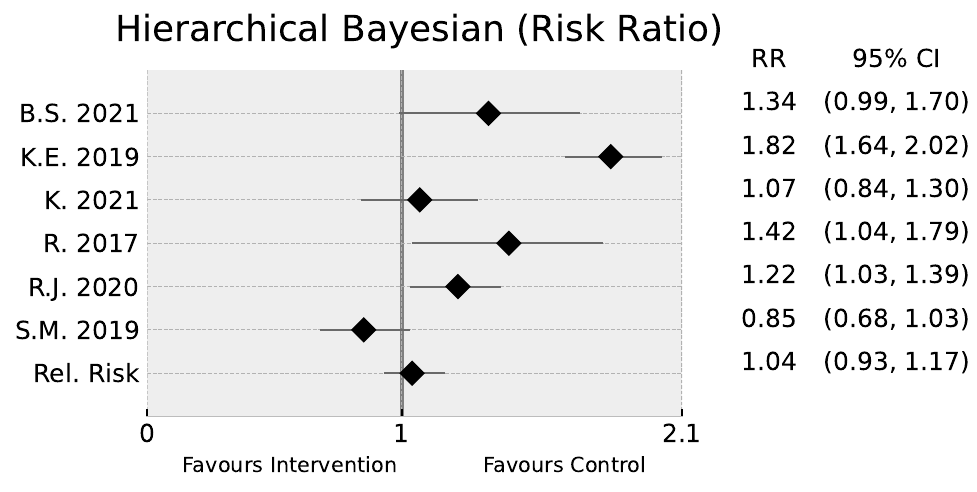}
    }
    \caption{An improved meta-analysis that accounts for the hidden variable of the Bishop score, which is used to determine whether mechanical dilation (intervention) or Pitocin (control) is used. This shows that current evidence is not sufficient to conclude there is any difference in the outcome of C-section rates. }
    \label{fig:bayesMetaResult}
\end{figure}

 Our Bayesian model in algorithm \ref{alg:generativeTolac} accounts for these factors and reports a risk ratio of 1.04 (CI 0.93-1.18). This indicates that there is no difference in risk of cesarean section when conditioned on physician preference to select for intervention based on Bishop scores (Fig. \ref{fig:bayesMetaResult}).

For TOLAC inductions, our results show that mechanical dilation is as effective and safe as Pitocin. This should alleviate hospital and provider concerns about the use of mechanical dilation to safely perform cervical ripening. This is important because in the case of a TOLAC, mechanical dilation is the only potentially available option for cervical ripening. 
In future studies the statistics of Bishop scores of each group should be reported for more robust conclusions and modeling.

\subsection{APGAR and Uterine Results}

Also of clinical interest in our meta-analysis is the rate of adverse uterine events (ruptures) and the APGAR\footnote{This is the common method of writing APGAR, though it is not an acronym --- it is named for its developer Virginia Apgar.} score is $<$ 7 by 5 minutes. Because the Bishop score is a direct measure of progress toward what Pitocin and mechanical dilation are attempting, the Bayesian model is necessary. The Bishop score does not have the same direct causal relationship with adverse uterine events or APGAR scores, and so a classic fixed-effect model is appropriate. Per our discussion in  \S \ref{sec:methods}, we use a Odds Ratio fixed-effect model congruent with standard practice. We note that applying our  algorithm \ref{alg:generativeTolac} to this data results in the same conclusion. 

\begin{figure}[!h]
    \centering
    \adjustbox{max width=\columnwidth}{%
    \includegraphics{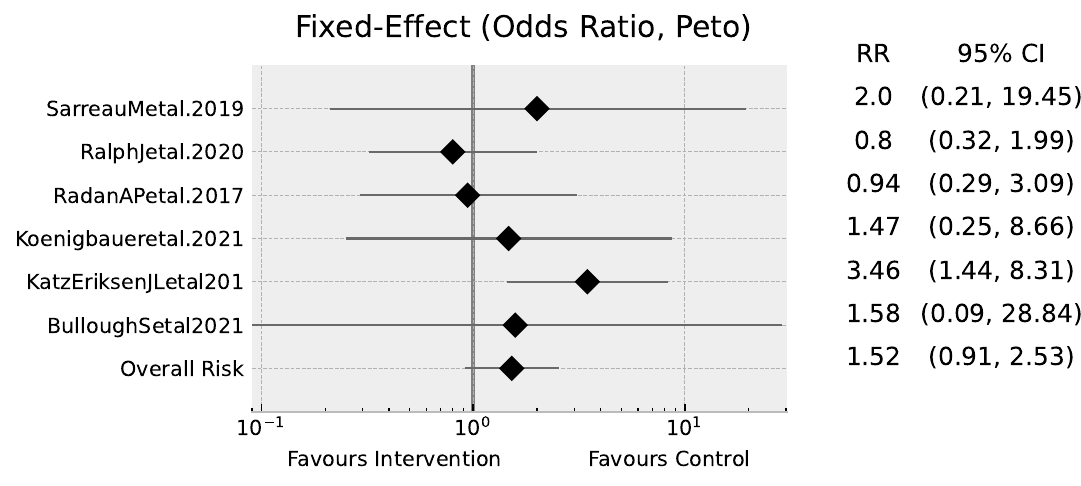}
    }
    \caption{A fixed-effect model is used due to the lack of a causal relationship between Bishop scores and adverse uterine events. The result is included for complete clinical relevance of our results. No significant effect between Pitocin and mechanical dilation is detected.}
    \label{fig:ForestPlotRupture}
\end{figure}

\begin{figure}[!h]
    \centering
    \adjustbox{max width=\columnwidth}{%
    \includegraphics{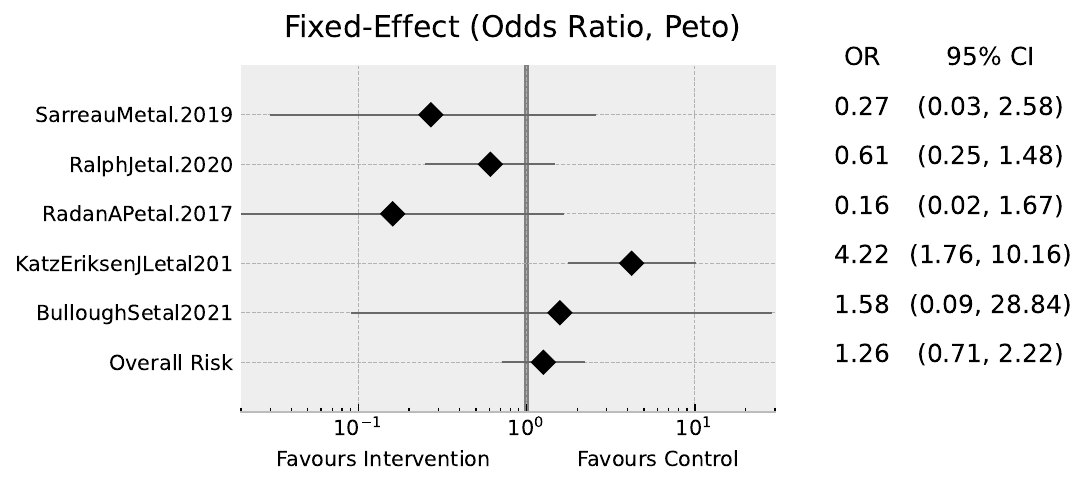}
    }
    \caption{Not all studies reported APGAR scores, and so only the relevant studies are included in the analysis. A fixed-effect model is used due to the lack of causal relationship and is included for the complete clinical relevance of our results. No significant effect between Pitocin and mechanical dilation is detected.}
    \label{fig:ForestPlotAPGAR}
\end{figure}

The mechanical dilation cohort has an adverse uterine event rate of 2.98\% (31 uterine rupture/dehiscence) and a 5-minute APGAR $<$7 rate of 1.92\%. The control cohort has an adverse uterine event rate of 1.73\% (52 uterine rupture/dehiscence) and an APGAR $<$7 at 5-minute rate of 1.83\%. The resulting odds indicate no significant difference for adverse uterine events (p = 0.136; CI 0.89-2.48
, Fig. \ref{fig:ForestPlotRupture}
) or APGAR $<$ 7 at 5min (p=0.434; CI 0.71-2.22
, Fig. \ref{fig:ForestPlotAPGAR}
).

\section{Discussion of Medical Context} \label{sec:discussion}

Management of induction of labor for TOLAC has an ever-increasing importance with the increasing rates of cesarean section. The risk of TOLAC failure and, more importantly, uterine dehiscence and rupture are of utmost concern when deciding induction methods. ACOG currently recommends avoidance of misoprostol due to the increased risk of uterine dehiscence and rupture, but there are limited recommendations regarding other methods of prostaglandins versus mechanical dilation with an unfavorable Bishop score. 

The current study compared complication rates and outcomes during TOLAC when using mechanical dilation. Overall, the results show no difference in the number of complications with both uterine and neonatal adverse events. When using the Bayesian model that also accounts for unreported Bishop scores, there was no statistical difference in VBAC rates between the mechanical dilation group and the control. 

	Like every meta-analysis, there are limitations. The main limitation is the small number of cohorts, which limits the power of this study. There is a paucity of high-quality literature describing the use of mechanical dilation in patients undergoing a TOLAC. This low number of studies gives us a cautionary pause. The risk of uterine rupture/dehiscence, while generally rare, is a catastrophic event. This leads to some hospitals and obstetricians not allowing induction of labor for patients wishing to undergo a TOLAC or limiting the induction agents used to Pitocin without allowing the use of mechanical dilation. Therefore, it is of utmost importance to determine if the use of mechanical dilation in patients undergoing TOLAC are at increased risk of complications. The current data does not support uterine scar as a contraindication for mechanical dilation for induction of labor. However, larger studies are warranted to provide additional evidence.

\section{Prior Methods In Confounded Meta-Analysis} \label{sec:confounded_prior}

The issue of statistical errors in medical work has long been known \citep{Glantz_1980}. Most prior works have focused on errors in the form of lack of assumption testing~\citep{Karadeniz_Uzabacı_Kuyuk_Kesin_Can_Seçil_Ercan_2019,Slutsky_2013,Hanif_Ajmal_2011} or of malfeasance or data entry errors \citep{Carlisle_2017,Ercan_Demirtas_2015}. To be clear, we are not accusing prior papers in the area of TOLAC in this article of these types of errors. Instead, our study is concerned with a type of error that physicians are not generally trained with regard to statistical study design: violations of the independently and identically distributed assumption via a shared causal factor~\citep{10722796}. In our case, the use of Pitocin vs. mechanical dilation is not independent because of the process that is used to decide between these factors. By accounting for this missing factor, we have determined there is no detectable statistical difference in outcomes. 

The issue of hidden variables impacting results is not unknown to the medical research community. The ``Many labs'' project explored this by replicating a study multiple times by different teams to attempt to quantify impediments to reproducibility and determined that unobserved hidden variables did have a measurable impact on results (but did not prevent replication in their case)~\citep{ManyLabs}. The key difference is our work considers a known unknown, whereas ManyLabs considers unknown unknowns. 
However, we are aware of no prior work in the medical literature that accounts for \textit{known but unobserved} hidden variables, for which their impact can be modeled under reasonable assumptions. While non-standard meta-analyses do occur, they are more likely in the form of a multivariate meta-analysis \citep{Gasparrini_Armstrong_Kenward_2012}, though we have found one biostatistics case of a similar hidden-variable modeling of known interactions ~\citep{Choi_Shen_Chinnaiyan_Ghosh_2007}.

Our meta-analysis is predicated on addressing a known mediating variable that was not accounted for in previous TOLAC studies. One previous line of work in confounded meta-analysis is to calculate an E-Value, which indicates the minimum effect size necessary to explain away a significant result~\citep{VanderWeele_Ding_2017}. If applied to our study we obtain a relatively small E-value of 2.13, which can be interpreted as saying the unaccounted impact of the Bishop score would need to be associated with at least a 2.13-fold reduction in relative risk across all studies to explain the apparent relation. This is not a statement of what is true, but a sensitivity analysis bounding the amount of unexplained effect that would need to exist. Given the Bishop score is a direct predictive and causal variable to the outcome of a TOLAC~\citep{doi:10.1080/01443615.2021.1916451}, we consider this E-value informative to the need for more careful design and study of Pitocin vs dilation for a TOLAC. 

While the E-Value is easy to apply, it does not inform a corrected estimate of the effect size we seek to study. In the review by ~\citep{doi:10.1146/annurev-publhealth-051920-114020}, other options that can be used to produce a corrected estimate are grouped into one-stage and two-stage. Broadly, both one and two-stage methods require expert or data-informed assumptions to be made about the size of the effect imposed by the confounding variable. 
\textit{These methods are most useful when the effect of the confounding variable can be meaningfully estimated}. There is no method for us to measure this effect absent first establishing the valid reason to pursue an RCT to obtain such data, thus the necessity of our approach. Though bounds have been developed for results under unobserved confounding~\citep{pmlr-v115-kilbertus20a,pmlr-v238-byun24a}, to wit, we are the first to look at meta-analysis. 
However, we are aware of no studies that estimate a marginal effect for the Bishop score, or more ideally, the individual components of a Bishop score, making the application of these techniques highly subjective. In contrast, our Bayesian model exploits the fact that the impact of the mediating factor is observable in the decision process of giving Pitocin or dilation, and allows us to more objectively specify hyper-priors based on known medical statistics instead of difficult-to-justify assumptions for one/two-stage corrections. This allows us to produce an answer with higher confidence. 

Our focus is on applying a machine learning technique, probabilistic programming, to develop a bespoke model for a clinically relevant question. This is to address a statistical issue in prior work and unreported information (the Bishop scores) that would affect the analysis. The machine-learning community is also not immune to these issues. Issues in machine learning replication have also been documented to stem from statistical errors ~\citep{Huang_2022,Bouthillier_Laurent_Vincent_2019} and insufficient information in published works ~\citep{Raff_2019,raff_what_2025,Raff2022a,raff_reproducibility_2023,Raff2020c}. In such cases, remediation of these issues is often only possible through full replication of the original study.
Other approaches to linear modeling, such as differential privacy~\cite{khanna_differentially_2025,khanna_sok_2024,swope_feature_2024,raff_scaling_2023,khanna_differentially_2023}, ordinal~\cite{Lu2022}, or sparsity-inducing models~\cite{lu_optimizing_2026,lu_high-dimensional_2024,Tibshirani1994}, may be relevant to medical applications but are beyond the current scope of work. 
As our work shows, the medical community can, in some cases, benefit from known causal structures that can allow the re-investigation of existing data. However, we still recommend that future studies collect and report Bishop scores to allow a more definitive conclusion. 

Last, we note that power analysis is confounded by the hidden variable itself. A study with low power is unlikely to detect an effect if one exists. The purpose of our study is not to try to maximize power, but to determine if a \textit{previously detected effect is actually supported by the data available}. From a clinical decision-making process, running the Bayesian meta-analysis is directly productive because if the study has an effect, then it supports the importance of the difference, irrespective of the missing variable. But if not, the ideal next action is to build evidence that includes the missing variable --- for which the meta-analysis's power analysis does not provide any benefit or change in the action taken. 

\section{Conclusion} \label{sec:conclusion}

To demonstrate the utility of our method, we have applied it to a real-world case study in OBGYN care of laboring mothers.  Working with physicians, we develop the generative story and specification of priors in a manner that is easy to adapt to other problems. Our case study is important as 
the rate of cesarean delivery has steadily increased, which, in turn, has led to an increase in TOLAC. Induction of labor in patients undergoing TOLAC is challenged by the limited agents available for cervical ripening. No significant difference was observed in the number of complications or in the rate of successful VBAC for patients undergoing TOLAC with the use of mechanical dilation compared to usual methods such as Pitocin. 
The results of this study are being used to help inform medical management of patients, provide the patients with increased options, and design Randomized Controlled Trials to further verify the results.

\bibliography{main}

\clearpage

\appendix

\section{Meta-Analysis Study Selection} \label{sec:meta_study_selection}

	A systematic literature search of clinical studies was performed to identify outcomes and clinical complications of mechanical dilation use during inductions of labor in patients undergoing TOLACs using the search engines MEDLINE (PubMed), Embase, Clinicaltrials.gov, and Ovid. The Preferred Reporting Items for Systematic Reviews and Meta-Analyses (PRISMA) guidelines were used to ensure the study was completed properly. Three OBGYNs independently performed the comprehensive search and evaluated all studies published 
 since January 1990
 using the following search terms: “TOLAC,” “balloon,” “VBAC,” “foley,” “mechanical dilation.” In total, 83 studies were identified.  

 \begin{figure}[!h]
     \centering
     \adjustbox{width=1\columnwidth}{%
     \includegraphics[]{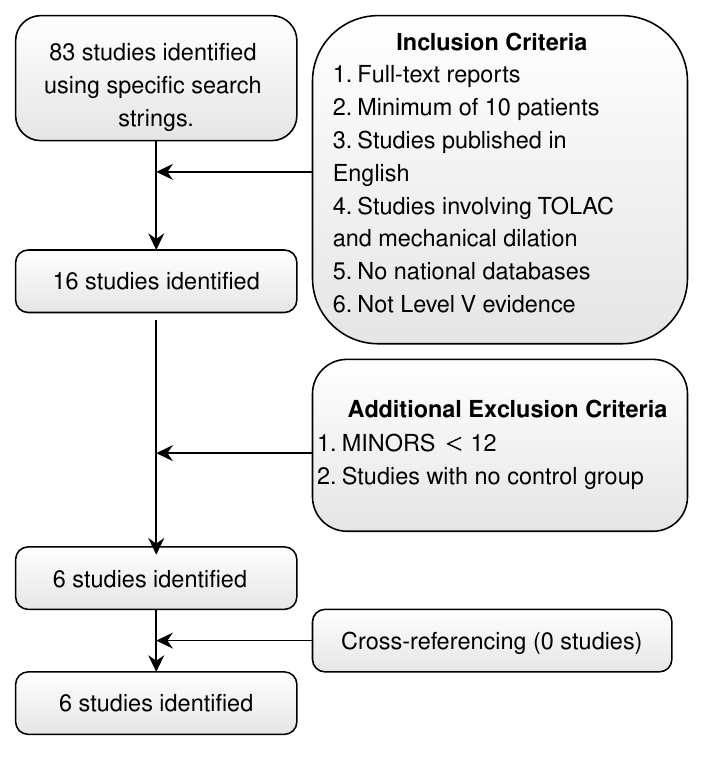}
     }
     \caption{Overview of the MINORS~\citep{moher2015a} process used to select the studies used for our meta-analysis. This is the standard accepted procedure in medical meta-analysis studies and so is used to ensure our results are clinically relevant. }
     \label{fig:paperIdentification}
 \end{figure}
 
	Each of the 83 studies was assessed for eligibility using the following inclusion criteria: (1) full-text reports; (2) minimum of 10 patients; (3) only studies published in English; (4) studies involving TOLAC and mechanical dilation; (5) no patient data from national databases; (6) not level V evidence (review articles, case reports). 
 
	A total of 16 studies and 7049 patients were identified that met the inclusion criteria. These were then read in full and cross-referenced to make sure all relevant studies were included. Cross-referencing identified no additional studies (Fig. \ref{fig:paperIdentification}). These studies were then designated a methodological index for non-randomized studies (MINORS) score~\citep{moher2015a}. No papers were excluded due to a MINORS score $<$ 12, and 8 studies were excluded due to lack of control groups.

Complications recorded in these studies included uterine rupture, uterine dehiscence, and APGAR scores at $\leq$5 minutes~\citep{hibbard2001a}. The rate of successful VBAC was also recorded. We investigated a potential difference in the number of complications and successful VBAC for patients using mechanical dilation during the induction process versus those who did not.

\section{Model Code}

The NumPyro model is given below in code form. The "I" variables in the input are "Intervention" and "C" is "Control". Similarly, "E" and "N" are for "Event" and "No-Event" respectively. Each is an array of the same length corresponding to a different study. 

\begin{minted}[linenos,tabsize=2,breaklines,fontsize=\footnotesize]{python}
def modelBishopHidden(IE, IN, CE, CN):
    log_eta = numpyro.sample("log(eta)", dist.Logistic(np.log(30), scale=1.0))
    mu = numpyro.sample("mu", dist.Beta(12,25))
    eta = jnp.exp(log_eta)
    tau = numpyro.sample("tau",  dist.HalfCauchy(np.sqrt(0.5/3)))
    tau_b = numpyro.sample("tau_B",  dist.HalfCauchy(np.sqrt(0.5/3)))
    intervention_prior = numpyro.sample("theta", dist.Normal(0, tau*tau))
    bishopImpact = numpyro.sample("Bishop Impact", dist.Normal(0, tau_b*tau_b))
    bishop_prior = numpyro.sample("Bishop", dist.Normal(0, 1))
    bishop_intervention_threshold_prior = numpyro.sample("Bishop Threshold delta", dist.Normal(0, 1))

    numpyro.deterministic("Total Relative Risk", expit(logit(mu) + intervention_prior)/mu)

    num_studies = IE.shape[0]
    with numpyro.plate("num_studies", num_studies):
      intervention_impact = numpyro.sample("Intervention Impact theta_i", dist.Normal(intervention_prior, tau*tau))
      bishopPopDist = dist.Normal(bishop_prior, 1)
      thresholdPop = numpyro.sample("Provider Discretion delta_i", dist.Normal(bishop_intervention_threshold_prior, 1))

      controlBishop = numpyro.sample("Control Bishop lambda_i", dist.LeftTruncatedDistribution(bishopPopDist, low=thresholdPop))
      catheterBishop = numpyro.sample("Catheter Bishop lambda_i", dist.RightTruncatedDistribution(bishopPopDist, high=thresholdPop))

      mu_control  = expit(logit(mu) + bishopImpact*controlBishop)
      mu_catheter = expit(logit(mu) + bishopImpact*catheterBishop + intervention_impact)
      prob_cs = numpyro.sample("P[CS | High Bishop]", dist.Beta(mu_control*eta,(1-mu_control)*eta))
      prob_cs_i = numpyro.sample("P[CS | Low Bishop]", dist.Beta(mu_catheter*eta,(1-mu_catheter)*eta))
      numpyro.deterministic("RR", prob_cs_i/prob_cs)

      IO = numpyro.sample("invervention_obs", dist.BinomialProbs(prob_cs_i, total_count=IE+IN), obs=IE)
      CO = numpyro.sample("control_obs", dist.BinomialProbs(prob_cs, total_count=CE+CN), obs=CE)
    return IO-CO
\end{minted}

\end{document}